\documentclass[review]{elsarticle}

\usepackage{hyperref}
\usepackage{graphicx}
\usepackage{booktabs}
\journal{Image and Vision Computing}

\usepackage{amsmath,amssymb} 
\usepackage{bm}
\usepackage{color}
\usepackage{graphicx}
\usepackage{subfig}
\usepackage{float}

\usepackage[linesnumbered,ruled]{algorithm2e}

\begin{document}

\begin{frontmatter}

\title{\textit{Co-Occurrence} of Deep Convolutional Features for Image Search}





\author{Juan I. Forcen}\corref{juan}
\address{das-Nano | Veridas,  31192, Tajonar, Spain\\ Dpt. Estad\'istica, Inform\'atica y Matem\'aticas \\ Institute of Smart Cities \\Universidad P\'ublica de Navarra \\ Campus Arrosad\'ia , 31006,  Pamplona, Spain}
\ead{jiforcen@das-nano.com}
\author{Miguel Pagola, Edurne Barrenechea}
\address{Dpt. Estad\'istica, Inform\'atica y Matem\'aticas \\ Institute of Smart Cities \\Universidad P\'ublica de Navarra \\ Campus Arrosad\'ia , 31006,  Pamplona, Spain}
\author{Humberto Bustince}
\address{Dpt. Estad\'istica, Inform\'atica y Matem\'aticas \\ Institute of Smart Cities \\ Universidad P\'ublica de Navarra \\ Campus Arrosad\'ia , 31006,  Pamplona, Spain \\
King Abdullazih Universitiy, Jeddah, Saudy Arabia}

\begin{abstract}
Image search can be tackled using deep features from pre trained Convolutional Neural Networks (CNN). The feature map from the last convolutional layer of a CNN encodes descriptive information from which a discriminative global descriptor can be obtained. We propose a new representation of \textit{co-occurrences} from deep convolutional features to extract additional relevant information from this last convolutional layer. Combining this \textit{co-occurrence} map with the feature map we achieve an improved image representation. We present two different methods to get the \textit{co-occurrence} representation, the first one based on direct aggregation of activations, and the second one, based on a trainable \textit{co-occurrence} representation. The image descriptors derived from our methodology improve the performance in very well-known image retrieval datasets as we prove in the experiments.
\end{abstract}

\begin{keyword}
\textit{co-occurrence} \sep image retrieval \sep feature aggregation \sep pooling
\end{keyword}

\end{frontmatter}

\section{Introduction}

Visual image search has rapidly evolved from variants of the bag-of-words model \cite{Lazebnik2006} based on local features, typically SIFT \cite{790410}, to approaches focused on deep Convolutional Neural Network (CNN) features \cite{Zheng2018}.  The first contributions in image retrieval using deep features were proposed by  Razavian et al. \cite{Razavian2014}  and  Babenko et al.  \cite{Yandex2015}. Basically they established different aggregation strategies for deep features and  demonstrated state-of-the-art performance in popular benchmarks. According to these results, taking into account that representations for image retrieval need to be compact,  i.e., around a few hundred dimensions, recent contributions have been made, in order to improve the quality of the final image representation. Relevant works as Tolias et. al. \cite{Tolias2015} or Kalantidis et al.  Basically in \cite{Kalantidis2016}, have focused in the methodology of feature extraction from the layers of the network into compact feature vectors using an off-the-shelf CNN, commonly known as a general feature extractor.  \cite{Kalantidis2016} established a  straightforward way of creating powerful image representations by means of multidimensional aggregation and weighting: an image is feed in a CNN, obtaining a tensor $A$, i.e. the activation maps of the last convolution layer, then these deep convolutional features  are aggregated to derive a final feature vector, i.e., the global image representation. Zheng et al. \cite{Zheng2018} define this category as {\it pre-trained single-pass} category of CNN-based approaches.

Other approaches  have  tried  to fine-tune the CNN with training datasets related to test datasets  \cite{Gordo2017}.  These approaches improve results in particular datasets, but have the drawback of requiring training datasets with expensive annotations depending on a category of each test set.

In computer vision has been widely used the concept of \textit{co-occurrence} matrix to represent textures among other visual features. A \textit{Co-occurrence} matrix \cite{4309314} is defined from an image being the distribution of co-occurring pixel values at a given spatial offset. Recently, this concept of \textit{co-occurrence} has been extended to the \textit{co-occurrence} of features activations in convolutional layers \cite{Shih2017}. In this approach, said \textit{co-occurrence} layer calculates the correlation between each pair of feature maps by means of the maximum product of the activations given a set of spatial offsets. This \textit{co-occurrence} representation obtain a 1-dimension vector which contains one correlation value for each possible pair of channels, therefore does not contain spatial information. Our proposed \textit{co-occurrences} representation is a tensor with the same dimension of the original activation tensor, which contains for each location their \textit{co-occurrences} information. In this paper we applied the \textit{co-occurrence} representation to obtain an improved image representation for image retrieval applications. The contributions of this paper can be summarized as follows:

\begin{itemize}

\item We propose a new definition for \textit{co-occurrence} representation of convolutional deep features.
\item We introduce a new concept of \textit{co-occurrence} filter. This filter is able to capture the dependencies between channels activations to obtain an improved  \textit{co-occurrence} representation.
\item We propose a linear and a bilinear pooling approach based on \textit{co-occurrence}, over off-the-shelf CNN convolutional features, for image retrieval.
\item We demonstrate in the experimental results  the effectiveness of our method to well-known image retrieval datasets, and we compare our method with the state-of-the-art techniques.
\end{itemize}

The rest of this paper is organized as follows. In Section \ref{sec:relatedwork}, related work with our proposal is recalled. In Section \ref{sec:cooc}, our new definition of \textit{co-occurrence} and its implementation by means of a \textit{co-occurrence} filter is proposed. Next, in Section \ref{sec:fixedcooc} are explained two pooling methodologies used to obtain a final image descriptor from our deep \textit{co-occurrences} tensor, followed by Section \ref{sec:learnablecooc} where a methodology to learn the best \textit{co-occurrence} representation is presented. Finally in Section \ref{sec:results_comparison} our method is compared with other \textit{co-occurrence} representation and with state of the art image retrieval methods. Finally, the conclusions and future work are highlighted in Section \ref{sec:conclusions}.

\section{Related work}\label{sec:relatedwork}

CNN-based retrieval methods have been emerged in recent years and are replacing the classical local detectors and descriptors methods. Several CNN models pretrained in giant datasets like Imagenet \cite{5206848}, serve as good choices for extracting features, including VGG \cite{Simonyan15} or  ResNet \cite{Krizhevsky2012}. Based on the transfer learning principle, the first idea was to extract an image descriptor from a fully-connected layer of the network, however, it has been observed  that the pooling layer after the last convolutional layer (e.g., pool5 in VGGNet), usually yields superior accuracy than the fully-connected descriptors and other convolutional layers \cite{Tolias2015}. Basically,  \cite{Razavian2014} and  \cite{Tolias2015} proposed a feature aggregation pipeline using max-pooling that, in combination with normalization and whitening, obtained state-of-the-art results for low dimensional image codes. 
Following these results, research efforts have been focused on the aggregation of the features from the pre-trained CNNs. This means, to identify proper spatial regions or weighting functions  to obtain a low dimensional image representation.  For example, Babenko et al. \cite{Yandex2015} used a global sum pooling with a center priority, and Kalantidis et al. \cite{Kalantidis2016} proposed a non-parametric spatial weighting method focusing on activation regions and a channel weighting related to activation sparsity with global sum pooling. Similarly, Cao et al. \cite{Cao2016} proposed a method to derive a set of base regions directly from the activations of the convolutional layer. Jimenez et al. \cite{Jimenez2017} studied the class activation maps  for spatial weighting and Mohedano et al. \cite{Mohedano2018} and Simeoni et al.\cite{Simeoni2018} proposed to use different human-based saliency measures for spatial pooling. However, all of these methods does not take into account the correlation between the features of the convolutional layers. This concept of correlation between image features was introduced by Yang et al. \cite{Yang2010}, named as feature  \textit{co-occurrence}, characterizes the spatial dependency of the visual features in a given image. Recently, Shih et al. \cite{Shih2017} introduce a new \textit{co-occurrence} representation for deep convolutional networks, demonstrating its effectiveness to exploit the information of visual features in the field of visual recognition. But, in the wrong side this method loses spatial information and its execution is very slow.

To overcome these problems, in this paper we consider that the correlation or interdependence between feature maps contains useful information, so in order to improve the final accuracy in the image search problem, we propose to add the \textit{co-occurrence} information to the final image descriptor using a new representation of deep \textit{co-occurrence}.

\section{Deep \textit{co-occurrence} Tensor of Deep Convolutional Features}\label{sec:cooc}

In convolutional neural networks when an image is feed in the CNN, the result after the last convolutional layer is an activation map $A$ of size $M \times N$ and $D$ channels, $A \in  \mathbb{R}^{M \times N \times D}$. This activations map represents how much is activated each feature (represented in channels) for a given spatial position.

The goal of \textit{co-occurrence} representation is to characterize the spatial dependency of the image features. Yang et al. \cite{Yang2010} call it \textit{``Spatial \textit{co-occurrence} Kernel"} and considered it as a count of the times that two visual features satisfy a spatial condition. Shih et al. \cite{Shih2017} present a new idea behind the \textit{co-occurrence} representation, recording the spatial correlation $c$ between a pair of feature maps $k$ and $w$, seeking the maximal correlation response for a set of spatial offsets $o_{ij} = [o_{ij,x},o_{ij,y}]^\top\in\mathbb{R}^2. i.e.$

\begin{equation}
c(k,w) = \max_{o_{ij}}\sum_{p\in[1,m]\times[1,n]}a_p^ka_{p+o_{ij}}^w
\end{equation}

where $a^k_p$ is the $k_{th}$ channel of $A$ at location $p$. And $a_{p+o_{ij}}^w$ is the $w_{th}$ channel of $a$ at location $p+o_{ij}$. 

In this approach the spatial information is lost in the resultant \textit{co-occurrence} vector $c$ of size $D^2$, due to the \textit{co-occurrence} between two channels is a single value. Moreover,  this high dimensionality makes it unaffordable in deep tensors like VGG with 512 channels. For this reason, Shih et al. \cite{Shih2017} add a $1\times 1\times N$ convolution filter to reduce the number of channels before the \textit{co-occurrence} layer. As a side effect this channel reduction causes a reduction of performance as demonstrated in \cite{Shih2017}; this representation was also used in  \cite{Elkerdawy2019}.

\subsection{Deep-Co-ocurrence-Tensor}\label{sec:tensor_cooc}

In this section we propose a new method to obtain a \textit{co-occurrence} representation from an activation tensor of a deep convolutional layer. This \textit{co-occurrence} representation is a tensor with equal dimensions than the activation tensor which encodes the correlation between feature maps for each tensor location.

\begin{figure}[h!]
\center
\includegraphics[width=0.45\textwidth]{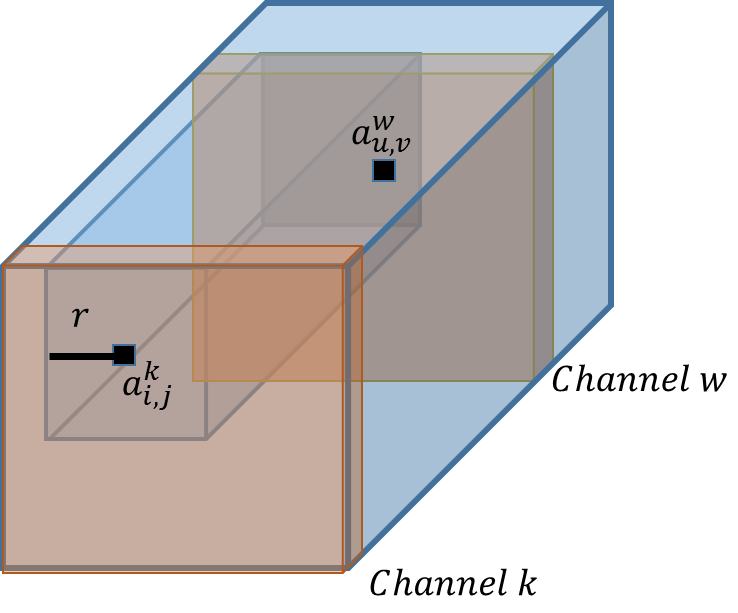}
\caption{A \textit{co-occurrence} occurs when the activation of two different activations $a_{i,j}^k$ and $a_{u,v}^w$ are greater than a threshold $t$ and both are spatially located inside of a given region.} \label{fig:COOC}
\end{figure}

We define that a \textit{co-occurrence} happens when the value of two different activations, $a_{i,j}^k$ and $a_{u,v}^w$, are greater than a threshold, $t$, and both are spatially located inside of a given region. (graphical interpretation is depicted in Figure \ref{fig:COOC}.). 

Given a convolutional map $A \in  \mathbb{R}^{M \times N \times D}$, containing a set of activations,  given a distance $r$ and a threshold $t$,  we define a positive \textit{co-occurrence} between two activations   $a_{i,j}^k$ and $a_{u,v}^w$ as: 
\begin{equation}
\rho(a_{i,j}^k,a_{u,v}^w) = \left\{ \begin{array}{ll} 1, & \text{if } \lvert i-u\rvert \leq r \text { and } \lvert{j-v} \rvert \leq r \text{ and} \\ & a_{i,j}^k > t \text { and }a_{u,v}^w > t \\
0, & \text{otherwise.}
\end{array} \right.
\end{equation}

The resultant elements of the \textit{co-occurrence} tensor are the sum of all the activations at the positive \textit{co-occurrences} divided by the number of channels. So, the \textit{co-occurrence} tensor is represented as $ C_T \in \mathbb{R}^{M\times N\times D}$, calculated by:

\begin{equation}
C_T(i,j,k) = \sum_{u=1}^{M} \sum_{v=1}^{N} \frac{1}{D-1}\sum_{w=1}^{D} \rho(a_{i,j}^k,a_{u,v}^w) \cdot  a_{u,v}^w
\label{eq:CT}
\end{equation}

Remark: The \textit{co-occurrence} of a channel with itself is not considered for the calculation of the \textit{co-occurrence}, for this reason all the activations aggregated are divided by $D-1$.

\subsubsection{\textit{Co-occurrence} and image representativeness}\label{sec:represent_cooc}
In order to study the representativeness of our proposed \textit{co-occurrence} tensor we visualize the pair-wise correlation of the query images of Oxford \cite{Philbin07} and Paris \cite{Philbin08} datasets. We calculate the total \textit{co-occurrence} vector $C_V \in \mathbb{R}^{1 \times D}$ as the sum of all the \textit{co-occurrences} per channel:

\begin{equation}
C_V(k) = \sum_{i=1}^{M} \sum_{j=1}^{N} C_T(i,j,k)
\end{equation}

We use these vectors $C_V$ of dimension $1\times D$ to compute the pair-wise correlation between images. The query-sets for both datasets contain 55 images in total, with 5 images of 11 classes of landmarks. We calculate the $C_T$ for each query image with $r=4$ and threshold $t$  as the average mean of all the activations at $A$ (being $A$ the last pooling layer called 'pool5' obtained  from a VGG16 pre-trained network in Imagenet dataset). 
 
\begin{figure}[!h]
  \centering
  \subfloat[Oxford]{\includegraphics[width=0.5\textwidth]{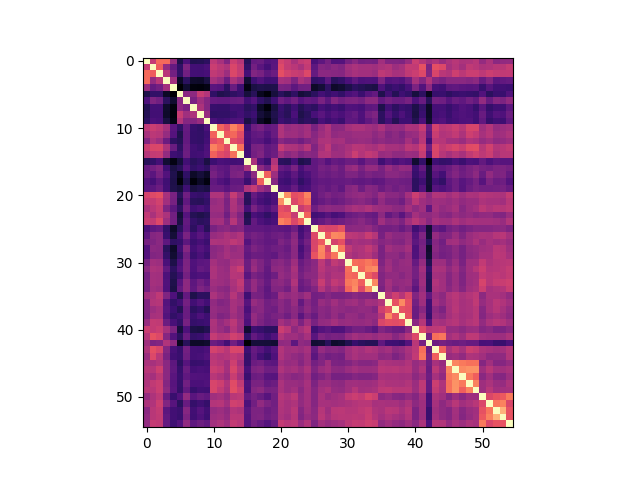}\label{fig:f1}}
  \hfill
  \subfloat[Paris]{\includegraphics[width=0.5\textwidth]{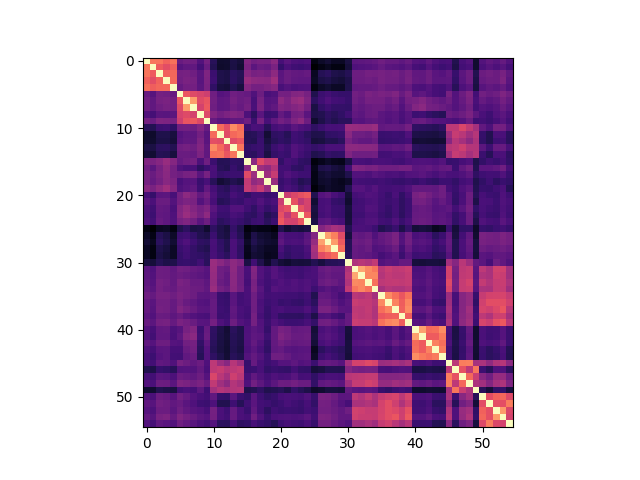}\label{fig:f2}}
  \caption{Correlation of \textit{co-occurrence} vector $CV$ for the 55 images in the query-set of Oxford (a) and Paris (b) datasets. Images are sorted by landmark.}
  \label{fig:correlation}
\end{figure}

In Figure \ref{fig:correlation} we observe that the \textit{co-occurrence} representation is highly correlated for images of the same landmark and less correlated for images of different landmarks. These figures, evidence that the \textit{co-occurrence} tensor contains discriminative information and therefore could be useful to obtain a representative vector applied to image search.

\subsubsection{\textit{Co-occurrence} filter representation}\label{sec:implementation_cooc}

An important advantage of our \textit{co-occurrence} representation is that it can be implemented using convolutional filters. We define the \textit{co-occurrence} filter as a convolutional filter: $F \in  \mathbb{R}^{D \times D \times S \times S} $, where $D$ is de number of channels in the activation tensor $A$, $S$ the window size being $S = 2 \cdot r + 1$, with $r$ the radius that defines the \textit{co-occurrence} region (see Figure \ref{fig:COOC}).

Note that activations do not compute to their channel \textit{co-occurrence} calculation. So all filters elements are initialized to one except the related with itself channel, that are initialized with zero or a small value $\varepsilon$, for example i.e. $1e-10$.

\begin{equation}
F_{a,b,c,d} \in  \mathbb{R}^{D \times D \times S \times S} = \left\{ \begin{array}{ll} 0, & \text{if } a=b\\
1. & \text{otherwise}
\end{array} \right.
\end{equation}

Given an activation tensor,  $A \in  \mathbb{R}^{M \times N \times D}$, of last convolution operator in a neural network,  the \textit{co-occurrence} tensor $C_T \in  \mathbb{R}^{M \times N \times D}$ can be obtained as a convolution between a thresholded  activation tensor and the \textit{co-occurrence} filter:
\begin{equation}
C_T = (A_{\rho_A} * F) \cdot \rho_A
\end{equation}

where $A_{\rho_A} = A \cdot \rho_A$ and $\rho_A \in  \mathbb{R}^{M \times N \times D}$, $\rho_A  = A > \overline{A} $, with $\overline{A}$ the average mean of the activation map, i.e.:
\begin{equation}
\rho_A(i,j,k) = \left\{ \begin{array}{ll} 1, & \text{if } a_{i,j}^k > \frac{1}{M\cdot N\cdot D} \sum_{i=1}^{M} \sum_{j=1}^{N} \sum_{k=1}^{D} a_{i,j}^k \\
0. & \text{otherwise}
\end{array} \right.
\end{equation}

\begin{figure}[!h]
  \includegraphics[width=0.95\textwidth]{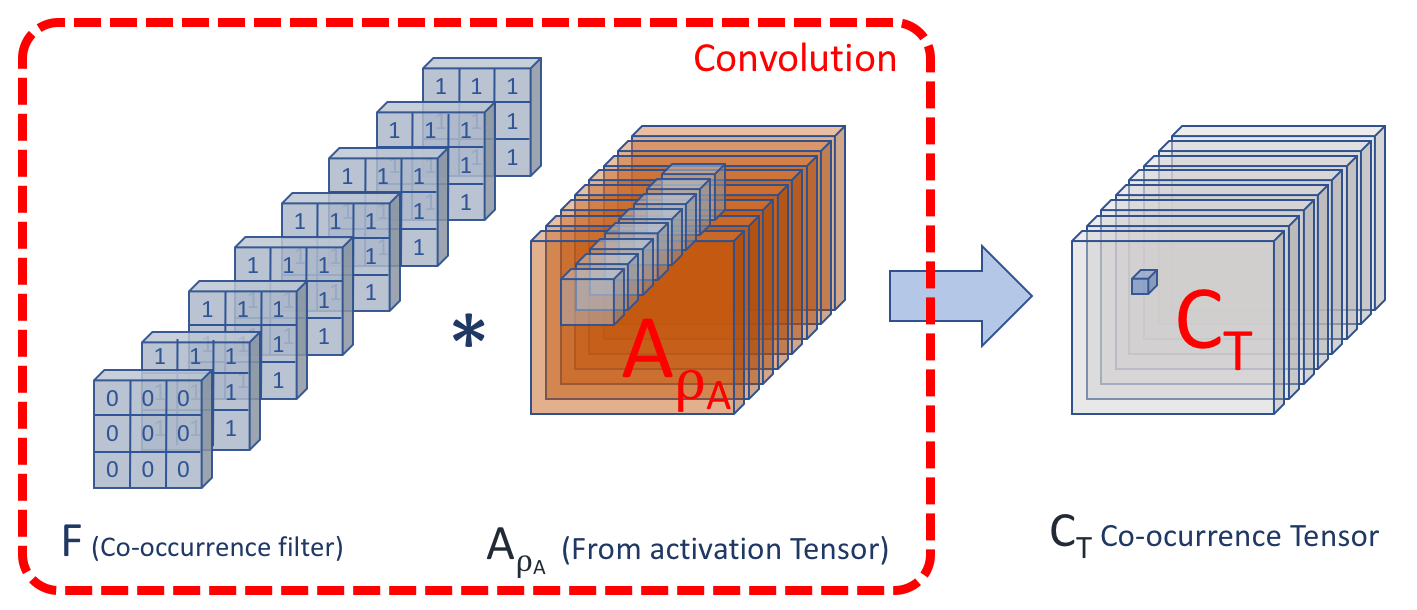}
  \caption{Example of the \textit{co-occurrences} filter $F_0 \in  \mathbb{R}^{D \times S \times S}$, convolved with $A_{\rho_A}$ to obtain a value in position $(i,j)$ in the \textit{co-occurrences} tensor (Best viewed in color).}
  \label{fig:learnable_filters_1}
\end{figure}

The pseudo-code of the implementation is shown in Algorithm \ref{fig:cooc_pseudocode} and in figure \ref{fig:learnable_filters_1} is depicted an illustration of the \textit{co-occurrence} tensor calculation.

\begin{minipage}{.8\linewidth}
  \centering

	\begin{algorithm}[H]
    	\footnotesize
    	\SetKwInOut{Input}{Input}
	    \SetKwInOut{Output}{Output}
	    \underline{calcCooc} $(A, S)$\;
    	\Input{$A$ : Tensor of activations with shape $D \times M \times N$ \\ $S$ : window size}
	    \Output{$C_T$ : Tensor of \textit{co-occurrences}}
	    $filters = ones(D, D, S, S)$ \\
	    $For \text{      } i=1 \text{      } to\text{      }  D:$ \\
	    $ \text{      }  \text{      }  \text{      }  \text{      } filters[i,i,:,:] = 1e$-$10$ \\
		$\rho_{A} = A > mean(A)$ \\
		$A_{\rho_{A}} = A \cdot \rho_{A}$ \\
		$C_T = conv2d(A_{\rho_{A}}, filters, padding=r) \text{ } / \text{ } (D-1)$  \\
		$C_T = C_T \cdot \rho_{A} $ \\
		$return \text{ } C_T$
	    \caption{ \footnotesize \textit{Co-occurrence} tensor.}
  		\par
	    \label{fig:cooc_pseudocode}    
	\end{algorithm}
\end{minipage}
\\

This  \textit{co-occurrence} implementation based on convolutions, makes possible to calculate the \textit{co-occurrence} representation with no channel reduction. Moreover, it is simple and straightforward to learn an improved \textit{co-occurrence} representation by adding a \textit{co-occurrence} layer in a trainable architecture.

In the next sections are introduced two different ways to use the \textit{co-occurrence} tensor applied to image retrieval. The first one, called \textit{Direct co-occurrences} uses a straightforward approach to calculate \textit{co-occurrence} (see Algorithm 1) and in the second one called \textit{Learnable co-occurrences} a  \textit{co-occurrence} filter is learned to obtain an improved \textit{co-occurrence} representation.

Implementation of the proposed method and some examples, in PyTorch are publicly available: \url{https://github.com/jiforcen/co-occurrence}.

\section{Direct \textit{co-occurrences}}\label{sec:fixedcooc}

Off-the-self methods of image retrieval obtain a compact representation of the images and queries to perform the query search, typically in three steps: (1) to feed the image in a pre-trained CNN to extract its activation tensor, (2) to apply a pooling function to obtain a compact representation and (3) ro apply l2-normalization and pca / whitenning to reduce the dimensionality and increment the discriminative power.

\begin{figure}
	\center
	\includegraphics[width=0.85\textwidth]{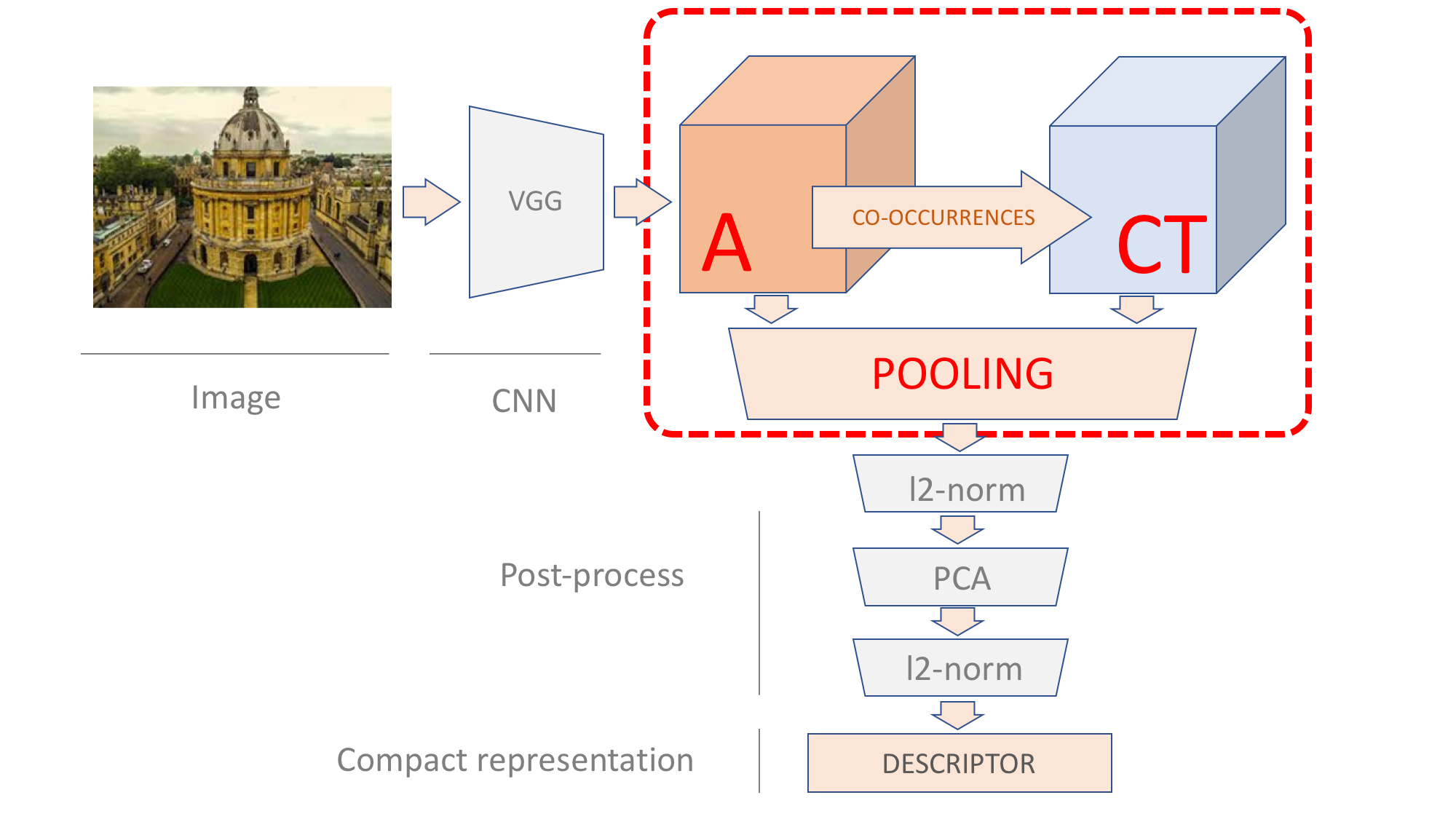}
	\caption{Aggregation pipeline to obtain the final compact image representation}
    \label{fig:aggregationpipeline}    
\end{figure}

In our method the compact representation, the second step,  is obtained by pooling the activation tensor with the \textit{co-occurrence} tensor (eee Figure \ref{fig:aggregationpipeline}). We have used two different methods, linear and bilinear pooling, to perform this aggregation and demonstrate the effectiveness of the \textit{co-occurrence} representation. Next, these two approaches are explained in detail.

\subsection{Linear weighted pooling}\label{sec:linear}

Following the lineal weighted pooling methodology proposed by Kalantidis et al. \cite{Kalantidis2016}, we transform the original tensor of activations into a new weighted tensor:
\begin{equation}
A_{i,j,k}' =  \alpha_{i,j}  \beta_k  A_{i,j,k}
\end{equation}
Where $ A \in \mathbb{R}^{M\times N\times D}$ is the tensor of activations from the last convolutional layer, $\alpha_{i,j}$ are the spatial \textit{co-occurrences} obtained from the \textit{co-occurrence} tensor and $\beta_k$ are channel \textit{co-occurrences}, also obtained from the \textit{co-occurrence} tensor. The final step of lineal pooling is the sum the new $A'$ tensor of activations by channel to obtain a single vector of dimension $1 \times D$.

The spatial \textit{co-occurrences} $\alpha_{i,j}$ basically are the normalized total \textit{co-occurrence} across all channels for every image location $(i,j)$, taking into account that spatial locations with large \textit{co-occurrences} across channels should correspond to discriminative locations of the image.

\begin{figure}[!h]
  \centering
 \begin{tabular}{cc}
  \includegraphics[width=0.45\textwidth]{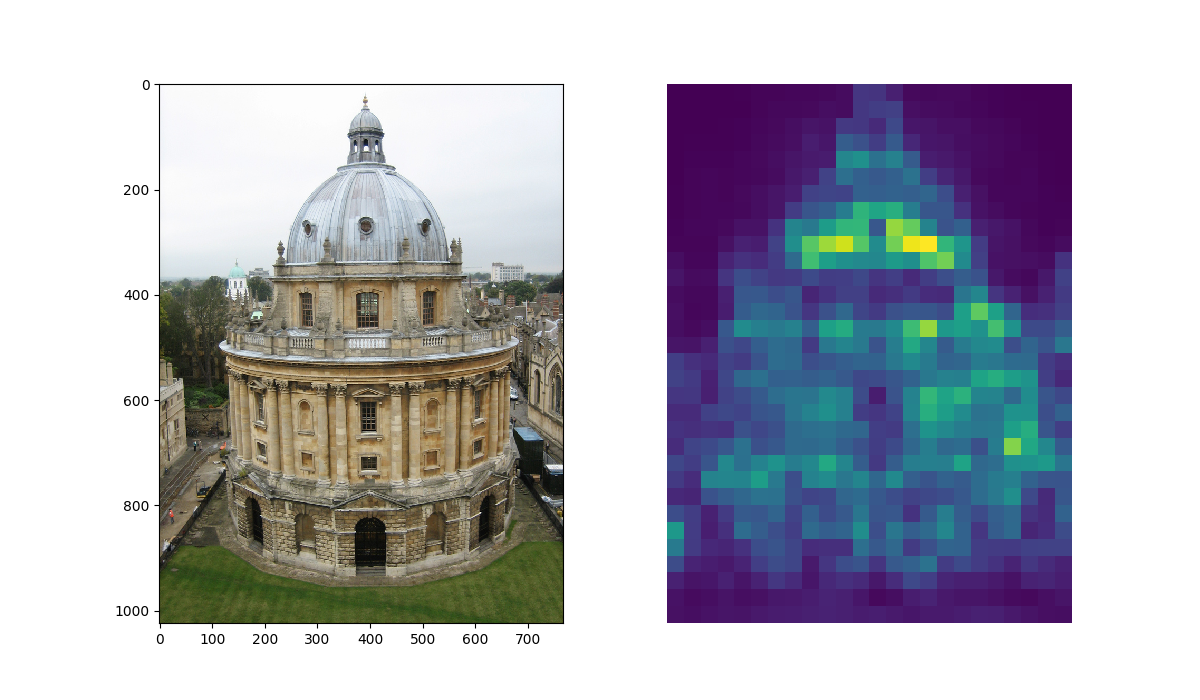} &
  \includegraphics[width=0.45\textwidth]{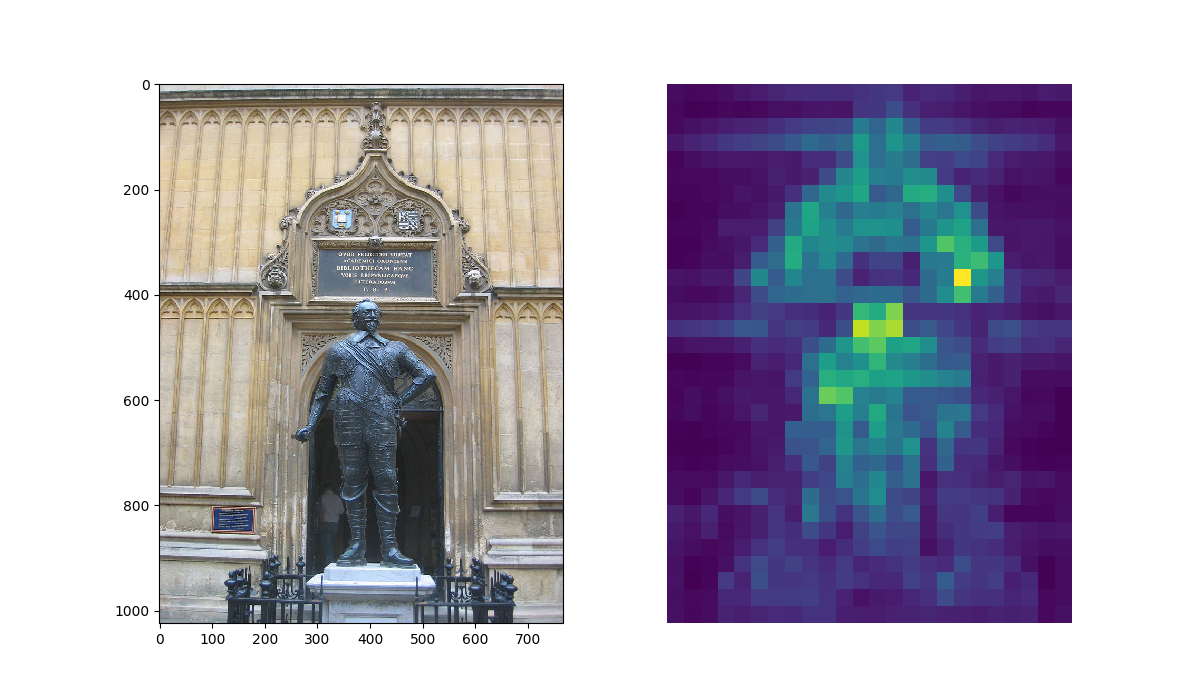} \\
  \includegraphics[width=0.45\textwidth]{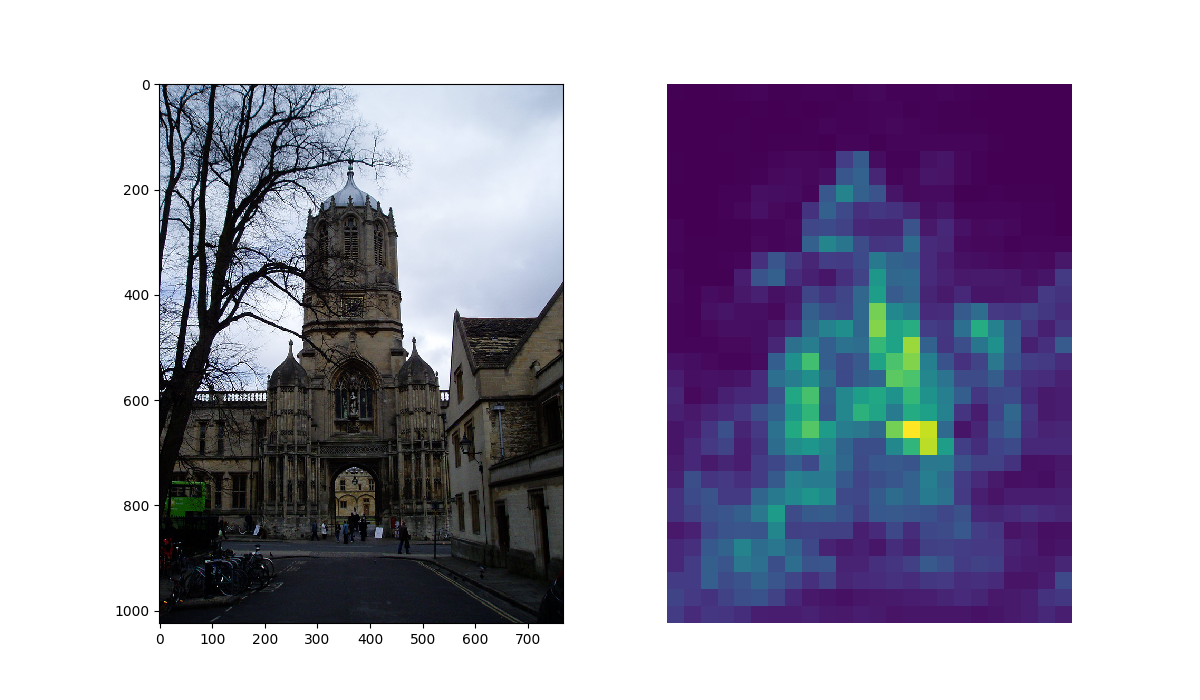} &
  \includegraphics[width=0.45\textwidth]{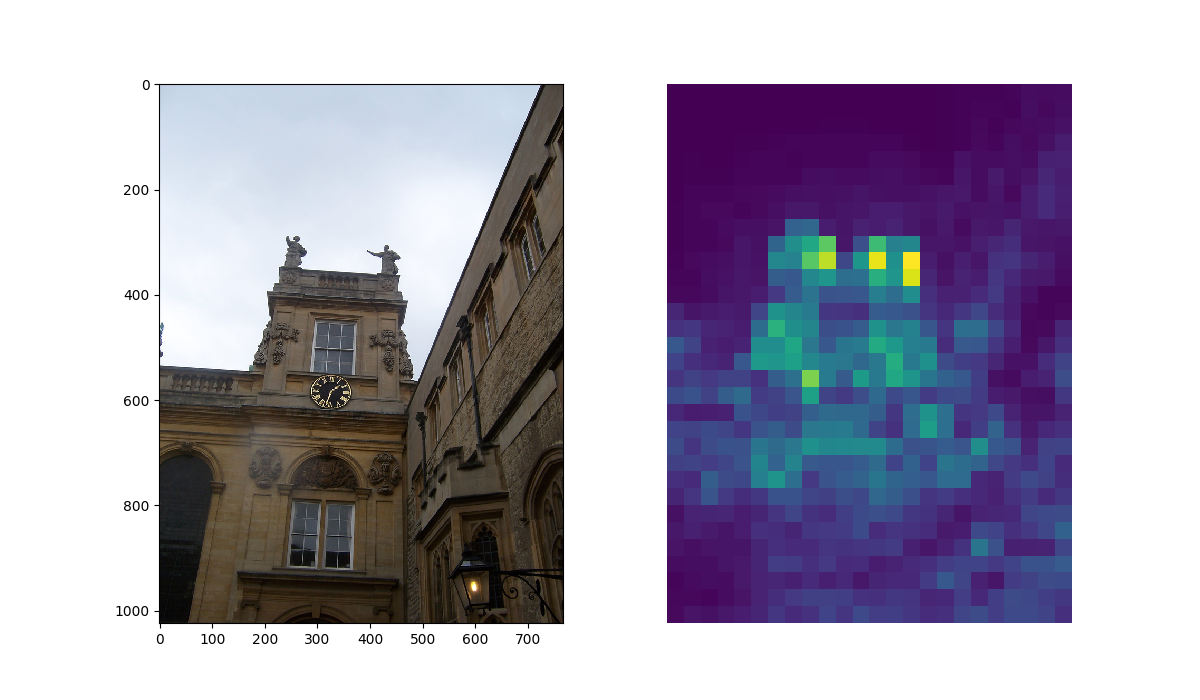}
 \end{tabular}
  \caption{Spatial \textit{co-occurrences} obtained from Eq. (\ref{eq:spatial}) for images of Oxford dataset.}
  \label{fig:Spatial_examples}
\end{figure}

We apply a power normalization to obtain the final spatial \textit{co-occurrences} matrix:
\begin{equation} 
\alpha_{i,j} = \left(  \frac{S(i,j)}{ \left( \sum_{i=0}^{i=M} \sum_{j=0}^{j=N} S(i,j)^a \right)^{1/a}} \right)^{1/b} \label{eq:spatial}
\end{equation}
where $ S \in \mathcal{R}^{M\times N}$ is the matrix of aggregated \textit{co-occurrences} from all channels per spatial location: 
\begin{equation}
S(i,j) = \sum_{k=1}^{D}CT(i,j,k) 
\end{equation} 
Figure \ref{fig:Spatial_examples} depicts spatial \textit{co-occurrences} $\alpha_{i,j}$ for several images of the Oxford dataset (\textit{co-occurrence} tensor $C_T$ is obtained by Equation \ref{eq:CT}). We visualize that our spatial \textit{co-occurrences} tends to give large values to locations with salient visual content and reject background information.

To calculate the importance of each channel the inverse of its \textit{co-occurrence} value is used, such a way we boost the contribution of rare features (in a similar way as term frequency inverse document frequency): 
\begin{equation}
\beta_k = log\left(  \frac{\sum_{l = 0}^{D}VC(l)}{\epsilon + VC(k)}\right)
\end{equation}
where $\epsilon$ is a constant to avoid division by zero.

\subsection{Bilinear pooling} \label{sec:bilinear}

Bilinear, or second order pooling, introduced by Tenenbaum et al. in \cite{Tenenbaum2000}, has the ability to capture pairwise correlations between channels of a descriptor. It have been successfully applied for different tasks, like semantic segmentation \cite{10.1007/978-3-642-33786-4_32}, fine grained visual recognition \cite{Guillot-Soulez2014} or face recognition \cite{7477593}. 
The bilinear pooling of a single dimension descriptor $x$ can be calculated as the outer product of $x$ an its transpose $x^T$. The outer product captures pairwise correlations between the feature channels and can model feature interactions. In our approach we combine the tensor of activations with the \textit{co-occurrence} tensor:
\begin{equation}
B(A,C_T) = \sum_{i}^M\sum_{j}^N A_{ij} \times C_{T i,j}^T
\end{equation}
Where $A_{i,j} \in {\Bbb R^{1 \times D}}$ is the vector of activations and $C_{T_{i,j}} \in {\Bbb R^{1 \times D}}$ the vector
of \textit{co-occurrences} at position $(i, j)$. The main disadvantage of bilinear pooling is the large size of the resultant
descriptor $B =  {\Bbb R^{D \times D}}$. For example an activation tensor with $D=512$  channels will produce a final descriptor with about $512 \times 512 \approx 250K$ values, which is excessive in most cases.
To tackle with that problem, Y. Gao et al. \cite{7780410} proposed compact bilinear pooling which is a kernelized view of bilinear pooling, achieving almost equal results of bilinear pooling with a final vector of 8K values which means a reduction of two orders of magnitude. We will use the compact bilinear pooling implementation in the experiments to reduce the $D\times D$ bilinear matrix into a  $1\times 8K$ vector.

\subsection{Experiments} \label{sec:results_fix}
In this section we present the results obtained with our proposal of direct {\it co-ocurrences} combined with linear and bilinear pooling. To perform the experiments we use common image retrieval datasets: Oxford \cite{Philbin07}, Paris \cite{Philbin08}, ROxford \cite{radenovic2018revisiting}, RParis \cite{radenovic2018revisiting} and Holidays \cite{Jegou:2008:HEW:1478392.1478419}. The Oxford Buildings Dataset consists of 5062 images with 11 different landmarks, each represented by 5 possible queries, Paris dataset is similar, created with 6412 Paris images. ROxford and RParis are a review of Oxford and Paris with three different difficulty levels. The Holidays dataset has 1491 images and 500 queries of holiday photos.

In addition we have used image retrieval common strategies, like spatial masks and query expansion techniques to reach improved performance results.

\subsubsection{Spatial masks} \label{sec:lin_sp}

Yandex et al. \cite{Yandex2015} supposed that the objects tend to be located close to the geometrical center of an image. So, they incorporate such centering prior with a simple heuristic, which assigns larger weights to the features from the center of the feature map. In the experimental study we evaluate this mask and we also try a new mask configuration, Top-Down, with larger weights to the features in the top of the feature map. This can be useful in datasets like Oxford and Paris which contains discriminant features in the top of the image such as high buildings and less discriminant objects in the bottom like floor, grass, trees, roads or even persons.

\subsubsection{Query expansion} \label{sec:lin_qe}

Query expansion repeats the search process based on a new query composed by the top ranked results on the first query \cite{6248018}, \cite{TOLIAS20143466}, \cite{Gordo2017}, \cite{Kalantidis2016}. We use two different query expansion methods in our experiments. The simplest method, called average query expansion {\tiny A}$QE$, is the average of the $n$ nearest descriptors after the first query. The second method, denoted alpha query expansion $\alpha QE$, presented by Radenovic et al. \cite{Radenovic2018}, which is a weighted aggregation of the nearest descriptors.

\subsubsection{Linear weighted pooling results} \label{sec:exp_lin}

For each image, the tensor of activations $A$ is the result of the last pooling layer "pool5" obtained  from a VGG16 pretrained network in Imagenet dataset (similar to \cite{Tolias2015}, \cite{Kalantidis2016}, \cite{Jimenez2017} and \cite{Mohedano2017}). The tensor $A$ has 512 channels, and its dimensions (height and width) are proportional to the input image size, due to we pass input images through the network with their original size. The tensor is processed following the scheme presented in Subsection \ref{sec:linear} resulting in a 512 dimensions single vector per image. In Oxford experiments, Paris dataset is used for PCA purposes, whilst in Paris and Holidays dataset Oxford dataset is used. For each image query, all of the images of the dataset are sorted according to their euclidean distance. Also a simple query expansion {\tiny A}$QE$ method is applied in Oxford and Paris dataset. To evaluate the performance in the given queries of these datasets we use the mean Average Precision metric ($mAP$), which is the standard procedure used in the literature.
 
In Table \ref{table:resultsE1} we present the comparison of our \textit{co-occurrence} based linear weighting pooling with other linear weighting methods. Ucrow \cite{Kalantidis2016} is the simplest way to aggregate the activation tensor in a compact descriptor, being calculated as the average of $A \in \mathbb{R}^{M\times N\times D}$ over the dimensions $M$ and $N$. Also, we compare with the method proposed in  Kalantidis et al. \cite{Kalantidis2016} called crow.

These two methods are compared with out method ChCO-$SC_{T}$ which is a combination of channel weighting $ChCO$ and spatial weighting \footnote{As in  \cite{Kalantidis2016} the parameters used in the power normalization were $a=2$ and $b=2$.} $SC_{T}$ to perform the linear weighted pooling aggregation of the activation tensor and \textit{co-occurrence} tensor.

Regarding the spatial masks (Subsection \ref{sec:lin_sp}) we have test a Top-Down weighting matrix, in which pixels of the top rows have higher weight than pixels in the bottom ($SpTD$) and a center prior weighting matrix \cite{Yandex2015}, in which pixels of the center have higher value than pixels in the borders ($SpCt$).

The \textit{co-occurrence} tensor $C_T$ was calculated with $r = 4$ \footnote{Previous experiments with $r = 4$, $r = 6$ and $r=8$ showed us that size influence was low, with the best case $r=4$.} and the threshold $t$ is the average mean of all the activations of the tensor $A$. 

Analysing Table 1, we can appreciate that the \textit{co-occurrence} based pooling is very helpful to obtain representative image vectors (ChCO-$SC_{T}$), because weighted vectors improve uniform aggregation (ucrow). Moreover, using ChCO-$SC_{T}$ our performance is similar in Paris and Oxford and better in Holidays than the crow\cite{Kalantidis2016} method, which is a state-of-the-art in pre-trained single pass methodologies. Therefore, the \textit{co-occurrence} tensor captures feature correlations and is able to provide better image representations. We have found that we get a large improvement in our results if the feature tensor $A$ is multiplied by the spatial masks, top-down $SpTD$ or center prior $SpCt$. Basically, top-down mask assumes upright images, which is the case of Oxford and Paris datasets due to all of the queries are buildings, and confirms that the spatial structure of the image is relevant in image retrieval.

\begin{table*}[h!]
\scriptsize
\begin{center}
\resizebox{\linewidth}{!}{%
\begin{tabular}{|l|c|c|c|c|c|c|c|c|c|}
\hline
\multicolumn{1}{|c|}{} & Oxford & \multicolumn{3}{c|}{ROxford} & Paris & \multicolumn{3}{c|}{RParis} & Holidays\\
\hline
& \multicolumn{1}{|c|}{} & \multicolumn{1}{c|}{Easy} & \multicolumn{1}{c|}{Medium} & \multicolumn{1}{c|}{Hard} & & \multicolumn{1}{c|}{Easy} & \multicolumn{1}{c|}{Medium} & \multicolumn{1}{c|}{Hard} & \\
\hline

Method &
mAP & mAP & mAP & mAP & mAP & mAP & mAP & mAP & mAP\\
\hline
\hline
ucrow &
66.0 & 60.53 & 41.15 & 11.98 & 75.8 & 74.75 & 57.69 & 30.20 & 81.1\\
crow\cite{Kalantidis2016} &
67.2 & 61.92 & 44.66 & 17.94 & 78.7 & 76.13 & 60.17 & 33.38 & 82.5 \\
ChCO-$SC_{T}$ &
67.05 & 61.31 & 44.52 & 18.71 & 79.17 & 77.16 & 60.69 & 33.89 & 83.22\\
ChCO-$SC_{T}$ + $SpTD$ &
\textbf{71.68} & \color{red}\textbf{63.37} & \color{red}\textbf{47.63} & \color{red}\textbf{21.96} & \textbf{80.89} & \textbf{78.79} & \textbf{62.59} & \textbf{36.84} &  83.04\\
ChCO-$SC_{T}$ + $SpCt$ &
66.82 & 61.67 & 43.05 & 14.91 & 80.18 & 77.32	& 61.05 & 33.79 & \color{red}\textbf{83.96}\\
\hline
\hline
ucrow + {\tiny A\par}$QE$ &
70.5 & 55.32 & 39.94 & 13.08 & 82.7 & 81.32 & 65.29 & 38.72 & - \\
crow + {\tiny A\par}$QE$ \cite{Kalantidis2016} &
71.5 & 55.75 & 42.11 & 17.89 & 85.5 & 81.82 & 67.93 & 43.16 & - \\

ChCO-$SC_{T}$ + {\tiny A\par}$QE$ &
71.30 & 56.41 & 42.46 & 18.93 & 85.31 & 83.59 & 69.03 & 44.12 & - \\
ChCO-$SC_{T}$ + $SpTD$ + {\tiny A\par}$QE$ &
\color{red}\textbf{77.48} & 60.27 & \textbf{46.58} & \textbf{21.19} & \color{red}\textbf{87.13} & 85.19 & \color{red}\textbf{71.01} & \color{red}\textbf{46.43} &  -\\
ChCO-$SC_{T}$ + $SpCt$ + {\tiny A\par}$QE$ &
72.99 & \textbf{61.83} & 44.57 & 16.79 & 86.08 & \color{red}\textbf{86.04} & 70.93 & 44.93 & - \\

\hline
\end{tabular}}
\caption{Results of linear weighted pooling aggregation based on the \textit{co-occurrence} tensor for the following datasets: Oxford, Paris, Holidays,  ROxford and RParis. (Coocurrences are calculated with r=4)}
\label{table:resultsE1}
\end{center}
\end{table*}

\subsubsection{Bilinear pooling results} \label{sec:exp_bilin}
In this experiment are evaluated the results obtained using the bilinear pooling method (Section \ref{sec:bilinear}). All parameters and measures are equal to the previous experiments (\ref{sec:exp_lin}). In Table \ref{table:resultsE2} we compare the results  of the bilinear pooling of activations $A$ and \textit{co-occurrences} $C_T$ ($BP(AC_T))$ with the performance of bilinear pooling with itself $A$ ($BP(AA)$). So, we can evaluate if the \textit{co-occurrence} provides useful information, also combined with the spatial masks. We conclude that the combination of \textit{co-occurrences} and activations is better in almost all the cases than the multiplication of the activations by itself ($BP(AA)$). Therefore, as in the previous experiment, we have proved that the \textit{co-occurrences} reflects the correlations of the features and provides better final image representations. The performance of the bilinear pooling is similar than the previous linear combination for Oxford and Paris datasets but worse in Holidays, and the Top-Down spatial mask influences in a similar way than the previous experiment.

\begin{table*}[h!]
\scriptsize
\begin{center}

\resizebox{\linewidth}{!}{%
\begin{tabular}{|l|c|c|c|c| c|c |c|c|c|}
\hline
\multicolumn{1}{|c|}{} & Oxford & \multicolumn{3}{c|}{ROxford} & Paris & \multicolumn{3}{c|}{RParis} & Holidays\\
\hline
& \multicolumn{1}{|c|}{} & \multicolumn{1}{c|}{Easy} & \multicolumn{1}{c|}{Medium} & \multicolumn{1}{c|}{Hard} & & \multicolumn{1}{c|}{Easy} & \multicolumn{1}{c|}{Medium} & \multicolumn{1}{c|}{Hard} & \\
\hline
Method &
mAP & mAP & mAP & mAP & mAP & mAP & mAP & mAP & mAP\\
\hline
\hline

$BP(AA)$ &
65.94  & 60.410 & 45.99 & 21.66 & 75.88 & 73.66 & 57.80 & 30.67 & 80.42 \\
$BP(AC_T)$	&
64.23 & 59.76 & 44.12 & 19.75 & 77.77 & 77.18 & 60.09 & 33.080 & \color{red}\textbf{82.62} \\
$BP(AC_T)$ + $SpTD$	 &
\textbf{71.00} & \color{red}\textbf{63.48} &  \color{red}\textbf{47.62} & \color{red}\textbf{22.83} & \textbf{79.56} & \textbf{77.72} & \textbf{62.14} & \textbf{37.09} & 79.51 \\

\hline
\hline
$BP(AA)$ + {\tiny A\par}$QE$ &
67.66 & 48.670 & 37.76 & 15.76 & 81.00 & 77.71 & 63.76 & 37.87 & - \\
$BP(AC_T)$ + {\tiny A\par}$QE$	&
66.08 & 51.24 & 38.22 & 15.9 & 82.97 & 83.03 & 67.06 & 40.85 & - \\
$BP(AC_T)$ + $SpTD$ + {\tiny A\par}$QE$	 &
\color{red}\textbf{77.25} & \textbf{56.01} & \textbf{43.69} & \textbf{19.33} & \color{red}\textbf{84.84} & \color{red}\textbf{83.17} & \color{red}\textbf{68.88} & \color{red}\textbf{44.38} & - \\

\hline

\end{tabular}}

\caption{Results of bilinear weighted pooling aggregation based on the \textit{co-occurrence} tensor for the following datasets: Oxford, Paris, Holidays,  ROxford and RParis. (Final vector size used is 512.)}
\label{table:resultsE2}
\end{center}
\end{table*}

Compact bilinear pooling aggregation of 512 depth vectors result in a vector of 8192 features, in  experiments showed in Table \ref{table:resultsE2} we have reduced said vector in the PCA step to 512 features, in order to compare it with linear weighted pooling. 

In Table \ref{table:resultsE21} are shown the results with larger final representation size. As we can observe larger vectors produce a great improvement in the $mAP$ results, being much better than the linear approach.

\begin{table*}[h!]
\scriptsize
\begin{center}
\resizebox{\linewidth}{!}{%
\begin{tabular}{|l|c|c|c|c|c| c|c |c|c|c|}
\hline
&\multicolumn{1}{|c|}{} & Oxford & \multicolumn{3}{c|}{ROxford} & Paris & \multicolumn{3}{c|}{RParis} & Holidays\\
\hline
&& \multicolumn{1}{|c|}{} & \multicolumn{1}{c|}{Easy} & \multicolumn{1}{c|}{Medium} & \multicolumn{1}{c|}{Hard} & & \multicolumn{1}{c|}{Easy} & \multicolumn{1}{c|}{Medium} & \multicolumn{1}{c|}{Hard} & \\
\hline
Method & FV Size &
mAP & mAP & mAP & mAP & mAP & mAP & mAP & mAP & mAP\\
\hline
\hline

$BP(AC_T)$ + $SpTD$ & 512 &
71.00 & 63.48 &  47.62 & 22.83 & 79.56 & 77.72 & 62.14 & 37.09 & 79.51\\
$BP(AC_T)$ + $SpTD$ & 1024 &
72.74 & 65.22 & 48.83 & 23.93 & 82.19 & 79.88 & 64.15 & 39.34 &  \\
$BP(AC_T)$ + $SpTD$ & 2048 &
73.76 & 65.98 &	50.14 &	26.14 & 83.08 & 81.46 & 65.49 &	40.76 & 81.10 \\
$BP(AC_T)$ + $SpTD$ & 4096 &
\textbf{76.56} & \color{red}\textbf{67.86} & \color{red}\textbf{51.92} & \color{red}\textbf{27.15} & \textbf{83.68} & \textbf{81.93} & \textbf{64.95} & \textbf{39.92}  & \color{red}\textbf{81.75} \\
\hline

$BP(AC_T)$ + $SpTD$ + {\tiny A\par}$QE$ & 512 &
77.25 & 56.01 & 43.69 & 19.33  &84.84 & 83.17 & 68.88 & 44.38  & - \\
$BP(AC_T)$ + $SpTD$ + {\tiny A\par}$QE$ & 1024 &
77.49 &  58.67 &	46.21 &	21.99  &  87.45&  85.40 &	71.28 &	47.24  & - \\
$BP(AC_T)$ + $SpTD$ + {\tiny A\par}$QE$ & 2048 &
79.03 &  59.38 &	46.68 &	22.19  & 88.65 &  86.94 &	72.99 &	49.75  & - \\ 
$BP(AC_T)$ + $SpTD$ + {\tiny A\par}$QE$ & 4096 &
\color{red}\textbf{81.71} & \textbf{64.33} & \textbf{51.28} & \textbf{26.64} &  \color{red}\textbf{89.16} &  \color{red}\textbf{88.11} & \color{red}\textbf{73.74} & \color{red}\textbf{50.23} & - \\

\hline

\end{tabular}}

\caption{Comparison of bilinear accuracy incrementing the final vector size.}
\label{table:resultsE21}
\end{center}
\end{table*}

\section{Learning \textit{co-occurrences}}\label{sec:learnablecooc}

In this section is explained how a \textit{co-occurrence} filter, $F$, can be learned within a trainable architecture to obtain a better \textit{co-occurrence}, $C_T$, representation. We have used a Siamese architecture due to it is known to produce highly discriminative embeddings \cite{1467314} \cite{Koch2015SiameseNN}, \cite{10.1007/978-3-319-48881-3_56}.

\subsection{Siamese learning} \label{sec:siamese}

\begin{figure}[!h]
  \centering
  \includegraphics[width=0.5\textwidth]{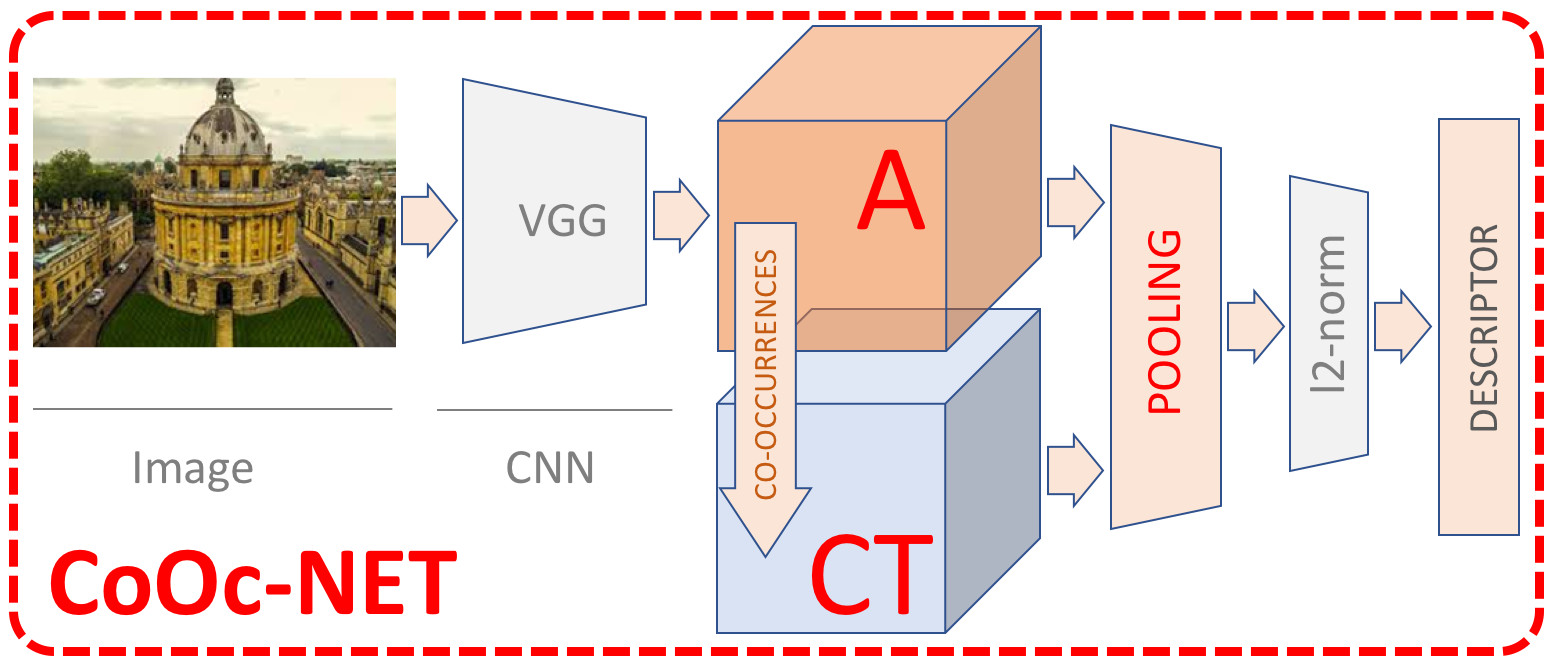}\\
  \includegraphics[width=0.5\textwidth]{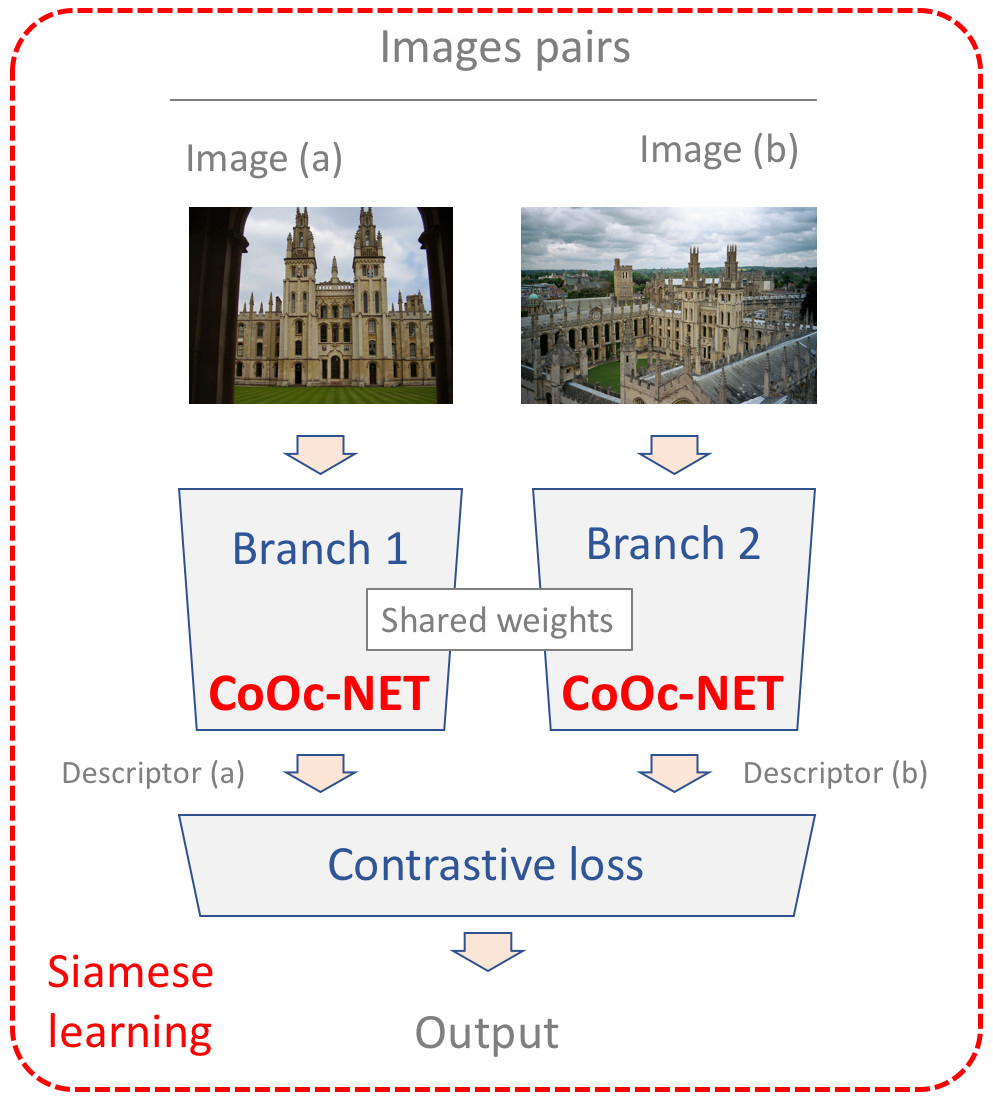}
  \caption{In the top the CoOcNET pipeline used to obtain a compact representation from an image combining the \textit{co-occurrences} tensor. In the bottom the siamese architecture based in two equal branches sharing weights and contrastive loss.}
  \label{fig:siamese_learning}
\end{figure}

The Siamese architecture is trained with paired samples, maximizing the similarity of positive pairs representation and also maximizing the dissimilarity of negative pairs. In Figure \ref{fig:siamese_learning} is presented the Siamese architecture adopted. It contains two equal branches that share parameters; each branch is represented as ”CoOc-NET”. The output of the ”CoOc-NET” is the l2-normalization of the resultant vector after the bilinear pooling operation between the activation tensor and its \textit{co-occurrence} tensor. In our experiments only the \textit{co-occurrence} filter is trained, and the rest of VGG layers are freeze, so the \textit{co-occurrence} filter weights change to obtain more discriminative representations.

This Siamese architecture is trained in combination with contrastive loss \cite{1467314} using as train input a pair of images $P_{Im} = [Im_a, Im_b]$ which produces descriptors $f_a$ and $f_b$. Label $Y(P_{Im})$ will be $1$ if the images of the pair corresponds to the same class or $0$ if not. Following this notation contrastive loss will be calculated as:

\begin{equation}
d=\|f_a - f_b\|_2 \\
\end{equation}

\begin{equation}
\L(P_{Im}) = Y(P_{Im}) * d^2 + (1 - Y(P_{Im})) * \max(\tau - d, 0)^2
\end{equation}

where $\tau$ is a parameter to establish a distance margin where dissimilar pairs influence the loss or not.

\subsection{Experiments} \label{sec:results_learnable}
In this section we present the results obtained using the trainable \textit{co-occurrence} filter, the experiments done are equivalent to section \ref{sec:results_fix}.

\subsubsection{Training Step}

In the training process of Siamese architecture we have used an image dataset called retrieval-SfM published by Radenovic et al. \cite{Radenovic2018}. This dataset contains 163k images grouped in 713 clusters of images. We have used the pairing images selection procedure of \cite{Radenovic2018}. Other parameters are: batch size equal to 5 and Adam \cite{Adam} optimizer with momentum 0.85 and learning rate 1e-9.

\begin{figure}[!h]
  \centering
  \subfloat[5 - 5]{\includegraphics[width=0.192\textwidth]{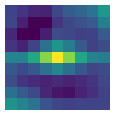}\label{fig:f1}}
  \subfloat[12 - 8]{\includegraphics[width=0.2\textwidth]{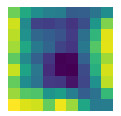}\label{fig:f2}}
  \subfloat[5 - 9]{\includegraphics[width=0.18\textwidth]{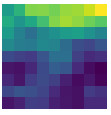}\label{fig:f3}}  
  \subfloat[1 - 9]{\includegraphics[width=0.193\textwidth]{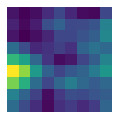}\label{fig:f4}}  
  \caption{Filter examples of size $9x9$ pixels where each spatial offset is weighted to obtain the best \textit{co-occurrence} representation. Each image represents the filter between a pair of channels. (Best viewed in color)}
  \label{fig:Spatial_examples_1}
\end{figure}

As we have explained previously only the \textit{co-occurrence} filter is learned, freezing the rest of the network. This \textit{co-occurrence} filter fine-tunning process takes only around 30 epochs to achieve the best result in the validation set. In  Figure \ref{fig:Spatial_examples_1} are shown four $9 \times 9$ filters after the training  process. Each filter represents the correlation between a pair of channels. 

In the Figure \ref{fig:Spatial_examples_1}, direct and learned \textit{co-occurrence} representations are compared. After the learning process the \textit{co-occurrence} representation shows the ability to put more emphasis in representative regions to discern between queries, for example buildings have higher activation values and a irrelevant objects or persons are dismissed from the \textit{co-occurrence} representation. In cases \ref{fig:Spatial_examples_1} (h) and \ref{fig:Spatial_examples_1} (i) we can see that the \textit{co-occurrence} between features representing humans are avoided, and only the \textit{co-occurrence} of the features representing buildings are taken into account.

\subsubsection{Evaluation}

In this section is evaluated the \textit{co-occurrence} representation after the \textit{co-occurrence} filter training process in a CoOcNET pipeline. The evaluation is performed similar to \cite{Radenovic2018}, with its same whitening procedure, alpha query expansion method $\alpha QE$, and testing also each query in multiscale, ms, ($1$, $\sqrt{\dfrac{1}{2}}$, $\dfrac{1}{2}$).

In table \ref{table:results_learnable_bilinear} are shown the results comparing bilinear pooling with the itself activation tensor $BP(AA)$, bilinear pooling combining the activation tensor and the \textit{co-occurrence} tensor with fix-weights $BP(AC_T)$ and with the learned weights $BP(AC_T)_{learn}$. It is easy to observe that learning  \textit{co-occurrences} $BP(AC_T)_{learn}$ obtains the best result, showing its ability to capture even better the relations between the features of the activation map.

\begin{figure}[!h]
  \centering
  \subfloat[]{\includegraphics[width=0.45\textwidth]{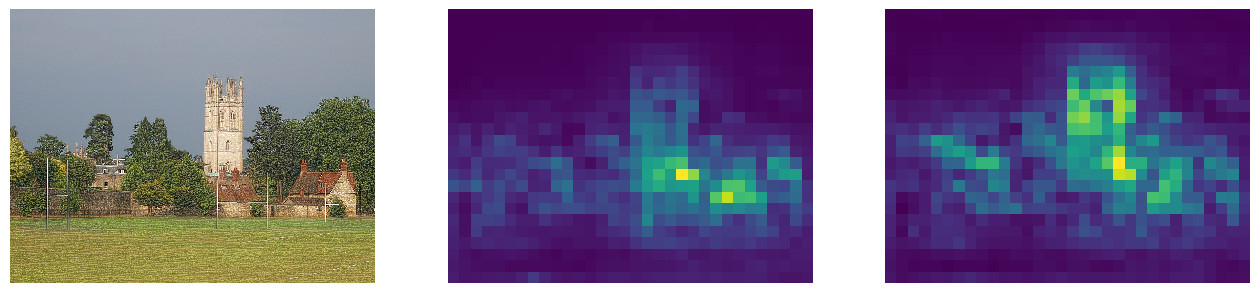}}
  \subfloat[]{\includegraphics[width=0.45\textwidth]{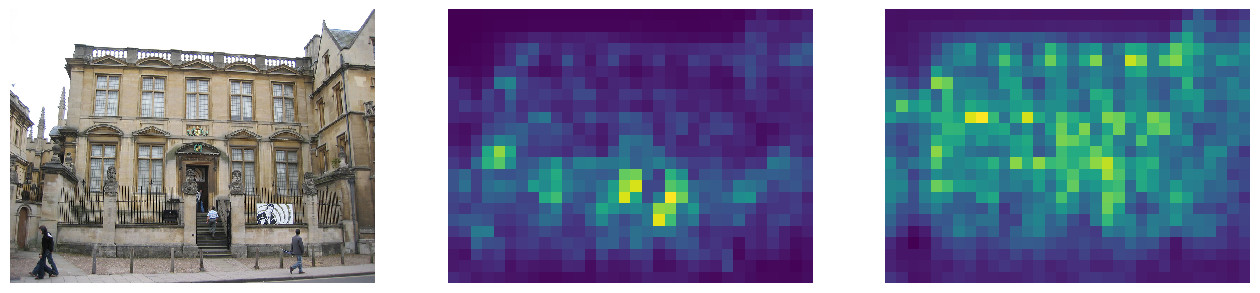}}
  \hfill
  
  \centering
  \subfloat[]{\includegraphics[width=0.45\textwidth]{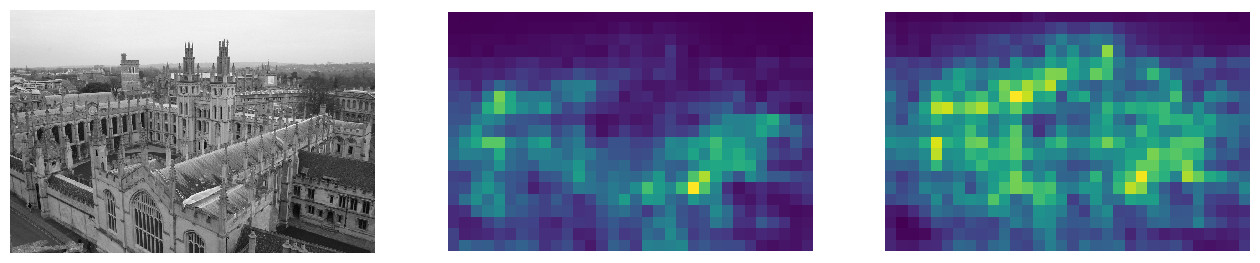}} 
  \subfloat[]{\includegraphics[width=0.45\textwidth]{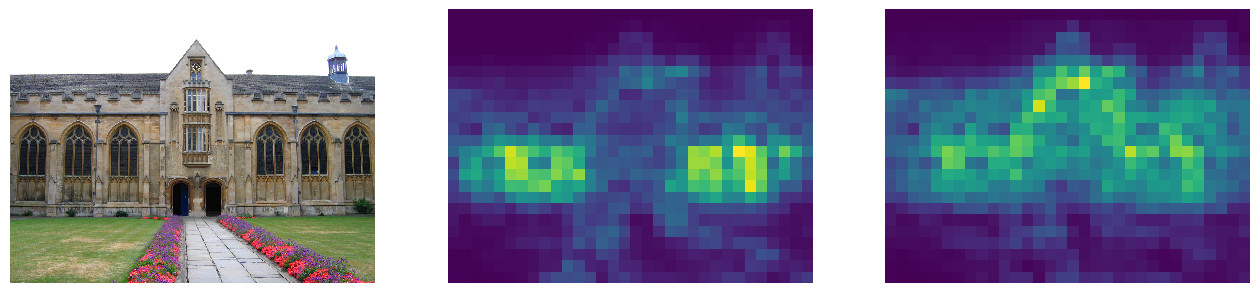}}  
  \hfill
  
  \centering
  \subfloat[]{\includegraphics[width=0.45\textwidth]{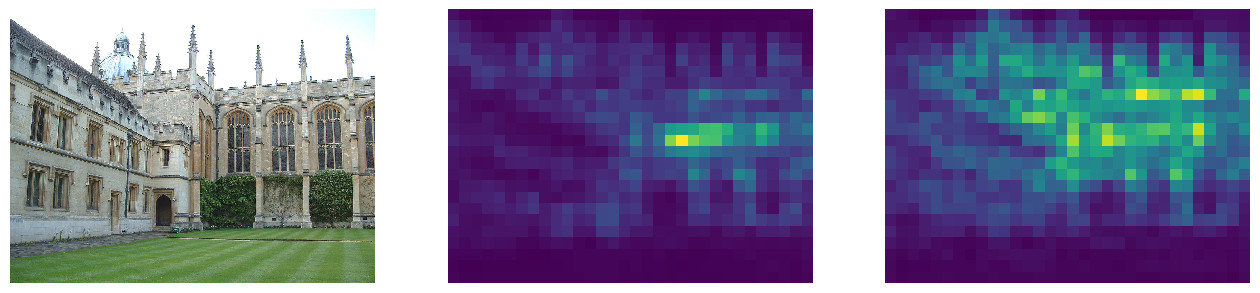}}
  \subfloat[]{\includegraphics[width=0.45\textwidth]{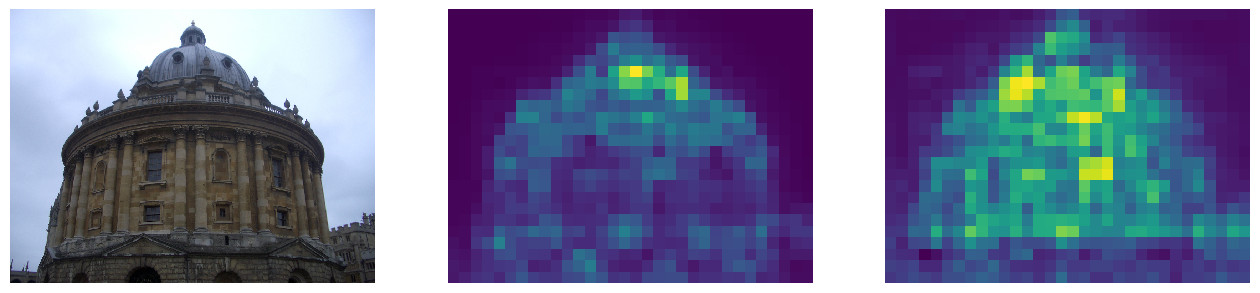}} 
  \hfill
  
  \centering
  \subfloat[]{\includegraphics[width=0.45\textwidth]{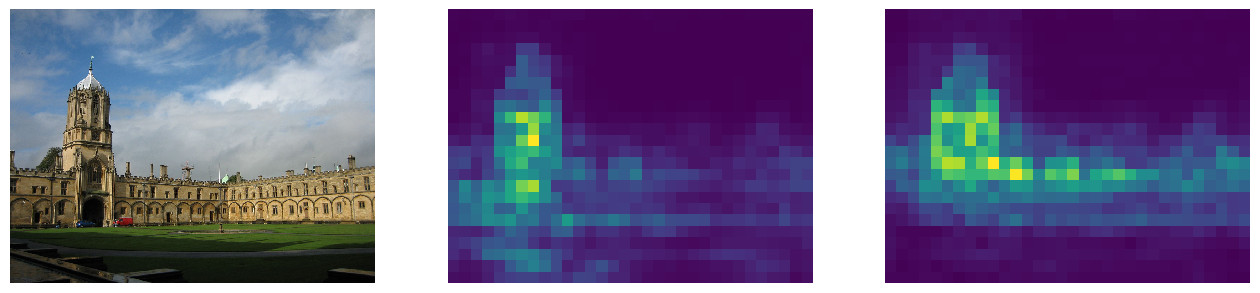}}  
  \subfloat[]{\includegraphics[width=0.45\textwidth]{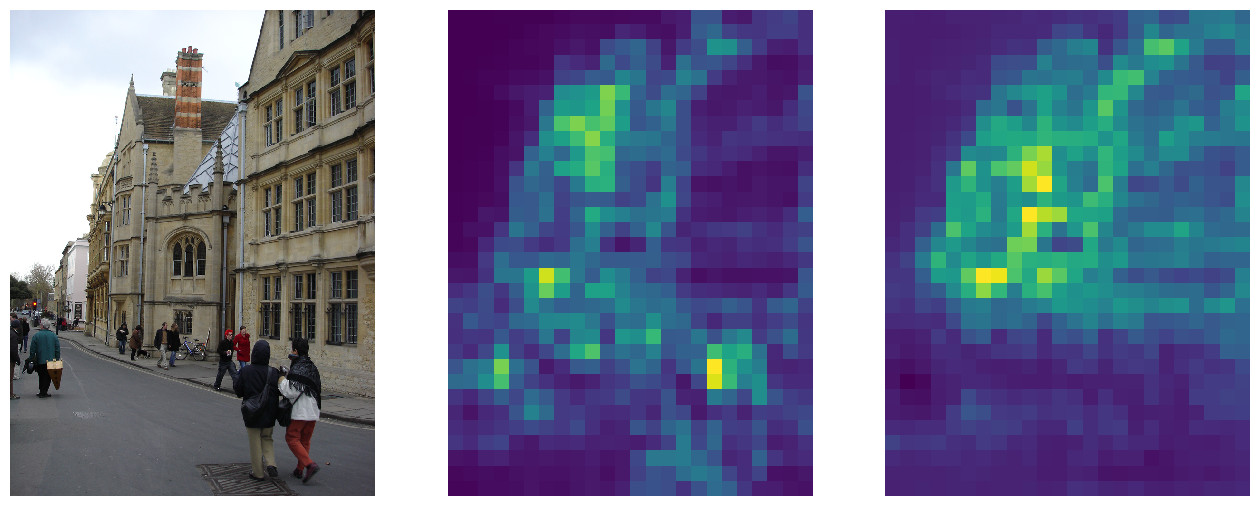}}  
  \hfill
  
  \centering
  \subfloat[]{\includegraphics[width=0.45\textwidth]{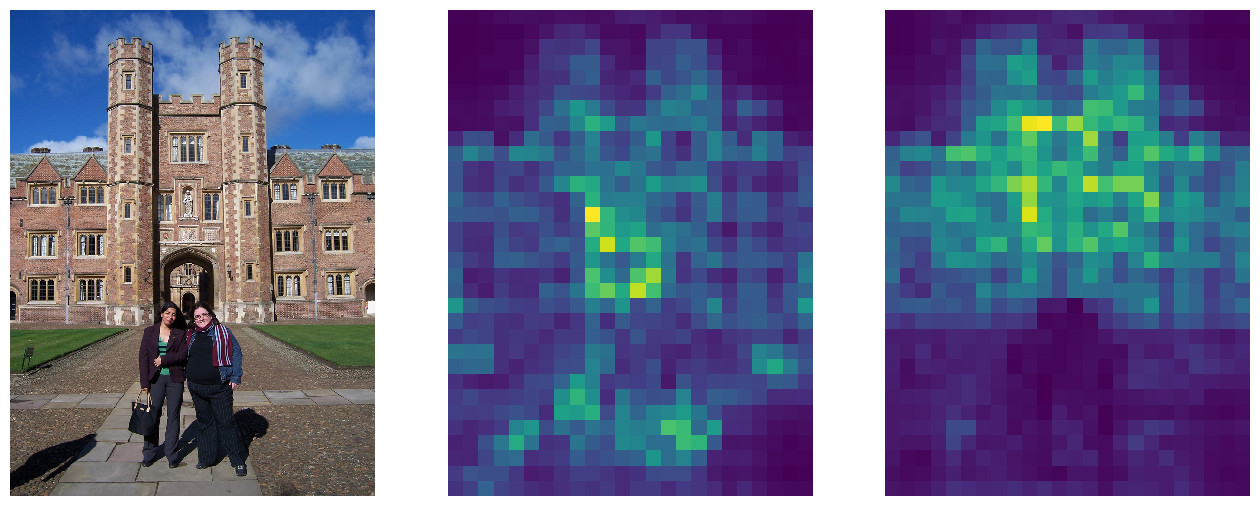}}
  \subfloat[]{\includegraphics[width=0.45\textwidth]{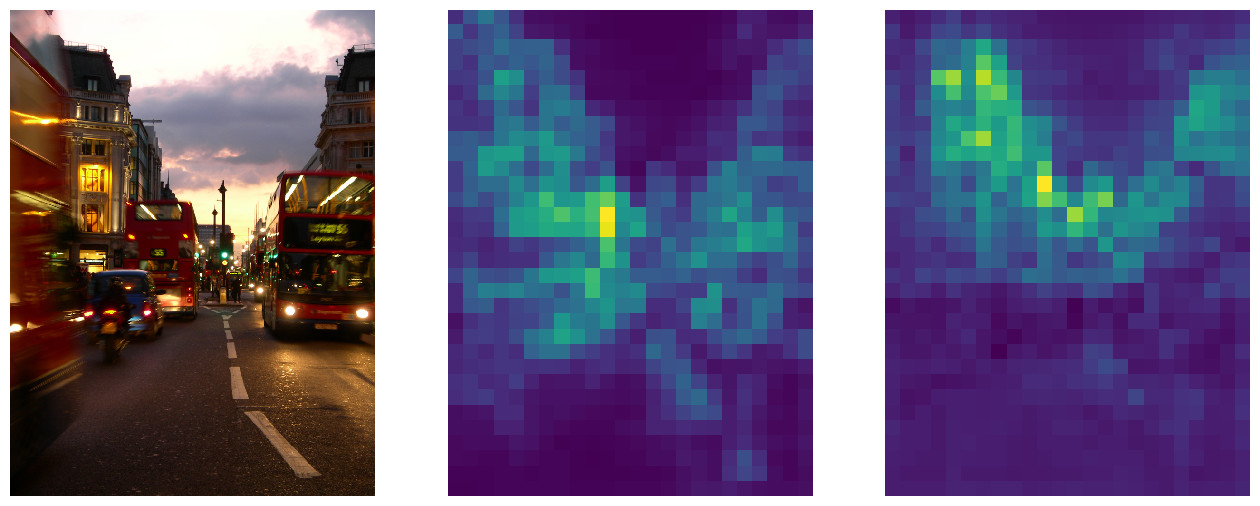}}      
  \hfill

  \caption{\textit{co-occurrence} representation of the \textit{co-occurrence} tensor (sum over channels) for some example images, showing for each one its \textit{co-occurrence} representation with fixed weights (left) and with trainable weights (right), (Best viewed in color).}
  \label{fig:learnable_filters}
\end{figure}

\begin{table*}[h!]
\scriptsize
\begin{center}
\resizebox{\linewidth}{!}{%
\begin{tabular}{|l|c|c|c|c| c|c |c|c|}
\hline
\multicolumn{1}{|c|}{} & Oxford & \multicolumn{3}{c|}{ROxford} & Paris & \multicolumn{3}{c|}{RParis}\\
\hline
& \multicolumn{1}{|c|}{} & \multicolumn{1}{c|}{Easy} & \multicolumn{1}{c|}{Medium} & \multicolumn{1}{c|}{Hard} & & \multicolumn{1}{c|}{Easy} & \multicolumn{1}{c|}{Medium} & \multicolumn{1}{c|}{Hard} \\
\hline
Method &
mAP & mAP & mAP & mAP & mAP & mAP & mAP & mAP\\
\hline
\hline

$BP(AA)$ + ms & 
76.67 & 72.97 & 55.67 & 31.24 &
90.31 & 89.69 & 72.22 & 48.0 \\
$BP(AC_T)$ + ms & 
76.79 & 72.53 & 56.34 & 32.93 &
89.55 & 90.13 & 71.86 & 48.22 \\
$BP(AC_T)_{learn}$ + ms & 
\textbf{80.75} & \textbf{74.65} & \textbf{57.94} & \textbf{33.68} &
\textbf{91.02} & \textbf{90.18} & \textbf{72.44} & \textbf{48.37} \\
\hline
$BP(AA)$ + $\alpha QE$ + ms & 
80.11 & 76.08 & 59.65 & 33.95 &
93.42 & 93.80 & 80.17 & 59.36 \\  
$BP(AC_T)$ + $\alpha QE$ + ms &
81.13 & 74.53 & 60.86 & 38.50 & 
93.23 & 93.93 & 80.35 & 60.44 \\
$BP(AC_T)_{learn}$ + $\alpha QE$ + ms &
\color{red}\textbf{85.76} & \color{red}\textbf{82.25} & \color{red}\textbf{67.04} & \color{red}\textbf{42.23} &
\color{red}\textbf{94.84} & \color{red}\textbf{94.21} & \color{red}\textbf{80.97} & \color{red}\textbf{61.42} \\
\hline
\end{tabular}}

\caption{Results of bilinear weighted pooling aggregation using the pipeline proposed in \ref{sec:learnablecooc} for learnable coocurrences in the following datasets: Oxford, Paris, Holidays,  ROxford and RParis. (Final vector size used is 8192.)}
\label{table:results_learnable_bilinear}
\end{center}
\end{table*}

\section{Comparison with State-of-the-art results}\label{sec:results_comparison}
In this section we compare the results of our method with the Shih et al. \cite{Shih2017} \textit{co-occurrence} interpretation method and also with other state-of-the-art methods in image retrieval.

\subsection{\textit{Co-occurrence} representation comparison}\label{sec:performance}

Shih et al. \cite{Shih2017} define \textit{co-occurrences} as the maximal correlation between a pair of feature maps, for a set of spatial offsets, whilst we define \textit{co-occurrences} as the sum of the activations inside a region, being the activations value above a threshold. (Section \ref{sec:tensor_cooc}).

In order to compare these two different interpretations, we have modified Shih et al. \cite{Shih2017} method to return a \textit{co-occurrence} tensor instead of a \textit{co-occurrences} vector. This modification consists in picking the summed maximum correlation map  instead of the own value. This produces a tensor 3D tensor $C_{T}^{'} = \mathbb{R}^{M \times N \times D^2}$ with $D^2$ channels. Each channel of this tensor represents the correlation between a pair of channels. The aggregation of all the correlations of each channel with the rest produce a tensor $C_{T}^{''} = \mathbb{R}^{M \times N \times D}$, with the same size than the original activation tensor $A$ as in our method.

In Table \ref{table:resultsE5} we present the comparison for linear and bilinear aggregation schemes using both \textit{co-occurrence} representation methods. The $mAP$ results obtained with the lineal aggregation are quite similar for both methods, but when bilinear aggregation is used our method outperforms significantly Shih et al.  method.
Furthermore Shih et al. method has an efficiency drawback in its implementation, because it  is necessary to find the maximum correlation value for each pair of channels.

\begin{table*}[h!]
\scriptsize
\begin{center}
\resizebox{\linewidth}{!}{%
\begin{tabular}{|l|c|c |c|c |c|c|c|}
\hline

\multicolumn{8}{|c|}{ROxford and RParis datasets} \\
\hline
\hline
\multicolumn{2}{|c|}{} & \multicolumn{3}{c|}{ROxford} & \multicolumn{3}{c|}{RParis} \\
\hline
\multicolumn{2}{|c|}{} & \multicolumn{1}{c|}{Easy} & \multicolumn{1}{c|}{Medium} & \multicolumn{1}{c|}{Hard}
& \multicolumn{1}{c|}{Easy} & \multicolumn{1}{c|}{Medium} & \multicolumn{1}{c|}{Hard} \\
\hline

Method & size & mAP & mAP & mAP & mAP & mAP & mAP \\
\hline
Shih ChCO-$SC_{T}$ + $SpTD$ & 512  &
64.3 & 47.64 & 20.86 & 79.45 & 62.92 & 37.15\\
ChCO-$SC_{T}$ + $SpTD$ & 512  &
64.28 & 47.02 & 20.30 & 79.01 & 62.18 & 35.79 \\
ChCO-$SC_{T}$ + $SpCt$ & 512 &
62.06 & 42.81 & 14.52 & 77.52 & 61.06 & 33.39 \\

Shih $BP(AC_T)$ + $SpTD$	 &  512 & 
56.27 & 44.51 & 21.94 & 73.15 & 61.25 & 37.95 \\
$BP(AC_T)$ + $SpTD$	 & 512 & 
63.48 &  47.62 & 22.83 & 77.72 & 62.14 & 37.09 \\

Shih $BP(AC_T)$ + $SpTD$	 & 4096 &
63.15 & 48.61 & 24.5 & 76.1 & 62.84 & 39.78 \\
 
$BP(AC_T)$ + $SpTD$	& 4096 &
\color{red}\textbf{67.86} & \color{red}\textbf{51.92} & \color{red}\textbf{27.15} & \color{red}\textbf{81.93} & \color{red}\textbf{64.95} & \color{red}\textbf{39.92} \\ 
 
\hline

\multicolumn{2}{|c|}{} & \multicolumn{1}{c|}{mAP (AQE)} & \multicolumn{1}{c|}{mAP (AQE)} & \multicolumn{1}{c|}{mAP (AQE)}
& \multicolumn{1}{c|}{mAP (AQE)} & \multicolumn{1}{c|}{mAP (AQE)} & \multicolumn{1}{c|}{mAP (AQE)} \\

\hline
Shih ChCO-$SC_{T}$ + $SpTD$ & 512  & 56.7 & 44.29 & 18.2 & 85.04 & 71.56 & 47.55 \\
ChCO-$SC_{T}$ + $SpTD$ & 512  &
57.98 & 44.32 & 16.56 & 84.57 & 70.37 & 45.45 \\
ChCO-$SC_{T}$ + $SpCt$ & 512  &
61.23 & 44.34 & 17.44 & 86.7 & 71.24 & 44.91 \\

Shih $BP(AC_T)$ + $SpTD$	 & 512 &
53.1 & 43.66 & 19.31 & 79.49 & 68.04 & 45.55 \\
$BP(AC_T)$ + $SpTD$ & 512 &
56.01 & 43.69 & 19.33 & 83.17 & 68.88 & 44.38 \\
Shih $BP(AC_T)$ + $SpTD$	& 4096 &
59.82 & 49.33 & 25.34 & 85.36 & 74.45 & 52.37 \\
$BP(AC_T)$ + $SpTD$ & 4096 &
\color{red}\textbf{64.33} & \color{red}\textbf{51.28} & \color{red}\textbf{26.64} & \color{red}\textbf{88.11} & \color{red}\textbf{73.74} & \color{red}\textbf{50.23}\\

\hline
\end{tabular}}
\caption{Results comparison betwen our proposed coocurrences and an adaptation os Shih et al method.}
\label{table:resultsE5}
\end{center}
\end{table*}

For this reason we have compared the execution time of the \textit{co-occurrence} tensor $C_T$ generation of both methods, with a subset of one hundred images of Paris6k. The feature map of each image was extracted with two pre-trained networks VGG \cite{Simonyan15} and  ResNet \cite{Krizhevsky2012}, resulting in tensors of 32x24 (width x height) with 512 channels (VGG) and 2048 channels (ResNet). Also, we have studied the performance with a smaller tensor of 32 channels depth (because is the depth of tensors used in the Shih et al.  implementation).

Table \ref{table:performance} shows the average time after fifty executions of each experiment using a computer with a CPU i7-7700K@4.20GHz, GPU GeForce GTX1080Ti, and 32GB of RAM.

\begin{table}[]
\begin{center}
\resizebox{0.8\linewidth}{!}{%
\begin{tabular}{@{}|l|l|l|l|l|@{}}
\hline
Tensor size        & ours (single) & DeepCooc\cite{Shih2017} (single) & ours (batch 5) & DeepCooc\cite{Shih2017} (batch 5) \\
\hline
32x24x512 (VGG)    & 0.977 ms  & 464.699 ms  & 0.695 ms & 495.66 ms  \\
32x24x2048 (ResNet) & 16.678 ms & 2417.895 ms & 5.578 ms & \textit{Out-of-memory} \\
32x24x32          & 0.258 ms  & 29.321 ms   & 0.318 ms & 35.227 ms  \\
\hline
\end{tabular}}
\caption{Performance comparison between \textit{co-occurrence} methods.}
\label{table:performance}
\end{center}

\end{table}

As we can see our implementation is more than hundred times faster than the Shih et al. method. Therefore, we have demonstrated that our method allows the use of \textit{co-occurrences} representations, breaking the performance barrier that made \textit{co-occurrences} calculation out of the reach for many applications.

\subsection{Comparison with State-of-the-art results}

In Table \ref{table:resultsEOA2}, we present the results in  ROxford and RParis datasets of state-of-the-art methods which uses VGG as feature extractor. In the pre-trained single pass category we improve the state-of-the-art performance with ChCO-$SC_{T}$ + $SpTD$ based in the linear aggregation of \textit{co-occurrences} against well known image retrieval methods like crow \cite{Kalantidis2016}, SPoC \cite{Yandex2015}, MAC and R-MAC \cite{Tolias2015} and GeM \cite{Radenovic2018}. Moreover, with bilinear pooling we can obtain a final vector representation with higher dimensions than the number of channels of the last VGG layer. Using Off-The-Self VGG and $BP(AC_T)$ with a final vector size of 8192 a great $mAP$ improvement is achieved.

Finally, we compare our \textit{co-occurrence} representation based on trainable \textit{co-occurrence} filter against GeM pooling \cite{Radenovic} using the same training and fine-tunning procedure for both methods. Again we demonstrate  a huge improvement as consequence of adding \textit{co-occurrence} information to the final vector representation, even when only the \textit{co-occurrence} filter is trained.

\begin{table*}[h!]
\scriptsize
\begin{center}
\resizebox{\linewidth}{!}{%
\begin{tabular}{|l|c|c |c|c |c|c|c|}
\hline

\multicolumn{2}{|c|}{} & \multicolumn{3}{c|}{ROxford} & \multicolumn{3}{c|}{RParis} \\
\hline
\multicolumn{2}{|c|}{} & \multicolumn{1}{c|}{Easy} & \multicolumn{1}{c|}{Medium} & \multicolumn{1}{c|}{Hard}
& \multicolumn{1}{c|}{Easy} & \multicolumn{1}{c|}{Medium} & \multicolumn{1}{c|}{Hard} \\
\hline

Method & size & mAP & mAP & mAP & mAP & mAP & mAP \\
\hline
\hline
\multicolumn{8}{|c|}{\textbf{Pre-trained single pass}} \\
\hline
\hline

ucrow & 512 &
60.53 & 41.15 & 11.98 & 74.75 & 57.69 & 30.20 \\
ucrow + {\tiny A\par}$QE$ & 512 &
55.32 & 39.94 & 13.08 & 81.32 & 65.29 & 38.72\\
crow \cite{Kalantidis2016} & 512 & 
61.92 & 44.66 & 17.94 & 76.13 & 60.17 & 33.38\\
crow + {\tiny A\par}$QE$ \cite{Kalantidis2016} & 512 &
55.75 & 42.11 & 17.89 & 81.82 & 67.93 & 43.16 \\ 

SPoC \cite{Yandex2015} & 512 & 
& 38.0 & 11.4 & & 59.8 & 32.4 \\
MAC  \cite{Tolias2015} & 512 & 
& 37.8 & 14.6 & & 59.2 & 35.9\\
R-MAC \cite{Tolias2015} & 512 & 
& 42.5 & 12.0 & & 66.2 & 40.9 \\
GeM \cite{Radenovic2018} & 512 & 
& 40.5 & 15.7 & & 63.2 & 38.8 \\

\hline

ChCO-$SC_{T}$ + $SpTD$ \textbf{(ours)} & 512  &
63.37 & 47.63 & 21.96 & 78.79 & 62.59 & 36.84 \\

ChCO-$SC_{T}$ + $SpTD$ + {\tiny A\par}$QE$ \textbf{(ours)} & 512  &
60.27 & 46.58 & 21.19 & 85.19 & 71.01 & 46.43 \\

$BP(AC_T)$  + ms \textbf{(ours)} &  8192 &
72.53 & 56.34 & 32.93
& 90.13 & 71.86 & 48.22 \\
$BP(AC_T)$ + $\alpha QE$  + ms \textbf{(ours)} & 8192 &
\textbf{74.53} & \textbf{60.86} & \textbf{38.50} & 
\textbf{93.93} & \textbf{80.35} & \textbf{60.44} \\
\hline
\hline
\multicolumn{8}{|c|}{ \textbf{Fine-Tunning}} \\
\hline
\hline
Radenovic VGG16-GeM \cite{Radenovic2018}  + ms & 512 &
& 61.9 & 33.7 & & 69.3 & 44.3 \\
Radenovic VGG16-GeM \cite{Radenovic2018} + $\alpha QE$  + ms & 512 &
& 66.6 & 38.9 &  & 74.0 & 51.0 \\

$BP(AC_T)_{learn}$ + ms \textbf{(ours)} &  8192 &
74.65 & 57.94 & 33.68 &
90.18 & 72.44 & 48.37 \\
$BP(AC_T)_{learn}$ + $\alpha QE$ + ms \textbf{(ours)} & 8192 &
\color{red}\textbf{82.25} & \color{red}\textbf{67.04} & \color{red}\textbf{42.23}
& \color{red}\textbf{94.21} & \color{red}\textbf{80.97} & \color{red}\textbf{61.42} \\

\hline
\end{tabular}}
\caption{Comparison with state-of-the-art results for ROxford and RParis datasets.}
\label{table:resultsEOA2}
\end{center}
\end{table*}

\section{Conclusions}\label{sec:conclusions}

In this work we have presented a new definition for \textit{co-occurrence} tensor of deep convolutional features. This \textit{co-occurrence} representation embeds relevant information of the image, allowing us to add discriminative information to the compact final image representations for image retrieval. In addition, our \textit{co-occurrence} implementation allows to learn the \textit{co-occurrence} filter to have better \textit{co-occurrence} representations.

In our approach we combine the proposed \textit{co-occurrence} tensor by means of weighted linear pooling and bilinear pooling with the original tensor of activations in a simple a straight forward pipeline. In the experimental results we have evidenced that the \textit{co-occurrence} tensor improve the results over the standard procedure, so the ability of \textit{co-occurrences} to capture additional information and create powerful image representations was demonstrated.
 
For future research, we plan to study other aggregation and normalization schemes for \textit{co-occurrence} and adapt our methodology on multi regional image representation \cite{Yandex2015}. 

\section*{References}

\bibliography{Deep_Cooc}
\end{document}